\documentclass[conference]{IEEEtran}
\IEEEoverridecommandlockouts
\usepackage{cite}
\usepackage{amsmath,amssymb,amsfonts}
\usepackage{algorithmic}
\usepackage{graphicx}
\usepackage{textcomp}
\usepackage{xcolor}
\usepackage{subfloat}
\usepackage{caption}
\usepackage{subcaption}
\usepackage{makecell}
\usepackage{float}
\usepackage{graphicx}
\usepackage{multicol}
\usepackage{multirow}
\def\BibTeX{{\rm B\kern-.05em{\sc i\kern-.025em b}\kern-.08em
    T\kern-.1667em\lower.7ex\hbox{E}\kern-.125emX}}
\begin{document}

\title{Logic-based AI for Interpretable Board Game Winner Prediction with
Tsetlin Machine
}

\author{\IEEEauthorblockN{Charul Giri}
\IEEEauthorblockA{\textit{Centre for AI Research} \\
\textit{University of Agder}\\
Grimstad, Norway \\
charul.giri@uia.no}
\and
\IEEEauthorblockN{Ole-Christoffer Granmo}
\IEEEauthorblockA{\textit{Centre for AI Research} \\
\textit{University of Agder}\\
Grimstad, Norway \\
ole.granmo@uia.no}
\and
\IEEEauthorblockN{Herke van Hoof}
\IEEEauthorblockA{\textit{Informatics Institute} \\
\textit{University of Amsterdam}\\
Amsterdam, the Netherlands \\
h.c.vanhoof@uva.nl}
\and
\IEEEauthorblockN{Christian D. Blakely}
\IEEEauthorblockA{\textit{AI and Real-Time Analytics} \\
\textit{PwC Switzerland}\\
Zurich, Switzerland\\
christian.blakely@ch.pwc.com}
}

\maketitle

\begin{abstract}
Hex is a turn-based two-player connection game with a high branching factor, making the game arbitrarily complex with increasing board sizes. As such, top-performing algorithms for playing Hex rely on accurate evaluation of board positions using neural networks. However,  the limited interpretability of neural networks is problematic when the user wants to understand the reasoning behind the predictions made.   In this paper, we propose to use propositional logic expressions to describe winning and losing board game positions, facilitating precise visual interpretation. We employ a Tsetlin Machine (TM) to learn these expressions from previously played games, describing where pieces must be located or not located for a board position to be strong. Extensive experiments on $6\times6$ boards compare our TM-based solution with popular machine learning algorithms like XGBoost, InterpretML, decision trees, and neural networks, considering various board configurations with $2$ to $22$ moves played. On average, the TM testing accuracy is $92.1\%$, outperforming all the other evaluated algorithms. We further demonstrate the global interpretation of the logical expressions, and map them down to particular board game configurations to investigate local interpretability. We believe the resulting interpretability establishes building blocks for accurate assistive AI and human-AI collaboration, also for more complex prediction tasks.

\end{abstract}

\begin{IEEEkeywords}
Tsetlin Machine, Winner Prediction, Interpretable AI, Hex
\end{IEEEkeywords}

\section{Introduction}

Board game winner prediction is a critical part  of state-of-the-art AIs for board game playing, such as AlphaZero~\cite{SilHub18General}. To guide a Monte-Carlo tree search towards a winning move, for instance, one needs to assess intermediate board configurations. Lately, state-of-the-art solutions use deep neural network architectures to evaluate board configurations and to propose which move sequences to explore. Such black-box algorithms achieve unsurpassed accuracy in board game position evaluation. However, their complex nature makes them hard to interpret as they do not reveal their internal decision mechanism. There are approximate post-hoc methods for explaining specific outputs such as LIME, yet, being approximate, they can be arbitrarily erroneous~\cite{Rudin2019}.

As AI integrates into our daily lives, interpretability becomes increasingly crucial. Whereas black-box methods can have significant ethical, legal, and social implications by slipping through biased and wrong decisions~\cite{kaur2021trustworthy}, algorithms that explain their output facilitate societal acceptance by being transparent~\cite{molnar2020interpretable}. An AI-powered recruitment tool developed by Amazon  was, for instance, found to be gender-biased and therefore scrapped \cite{dastin2018amazon}.

Lack of interpretation is problematic because it hinders quality assurance and human-computer collaboration. In contrast, an interpretable system enables human assessment of how the AI reasons, for overseeing the AI decision process. If the AI, in addition, can provide transparent decision alternatives, it becomes possible to augment AI-based decision-making with human intelligence, through collaboration. Indeed, humans and AI collaborating may be more capable than each alone. As exemplified by Hipp et al. \cite{hipp2011computer}, the combination of the tactical superiority of computers and the strategical superiority of humans led to the victory of amateur players assisted by AI in Cyborg Chess\footnote{Cyborg Chess, also called Advanced Chess, is a form of chess where the human player is assisted by a computer chess program to explore potential moves more effectively.}, facing grandmasters and supercomputers.

The above findings are in contrast to the traditional view that complex black-box models are necessary to maximize accuracy, creating a trade-off between interpretability and accuracy.  Further, Rudin suggests that interpretable models can perform competitively if the representation of the problem is adequate \cite{Rudin2019}.

Finding competitive interpretable representations for complex board games like Hex is an unresolved challenge. Contemporary interpretable techniques, like decision trees and XGBoost, employ greedy learning strategies or rely on model simplifications. They, therefore, struggle with learning complex patterns such as board configurations that lead to a win or loss.

In this paper, we propose a competitive interpretable representation for the board game Hex, learned by a Tsetlin Machine (TM)~\cite{articleTM}. TM is a recent technique for pattern recognition that uses a team of Tsetlin automata \cite{Tsetlin1961} to learn patterns expressed in propositional logic. TMs have obtained competitive accuracy, memory footprint, and learning speed on several benchmark datasets~\cite{abeyrathna2021massively}. They have been particularly successful in natural language processing, including explainable aspect-based sentiment analysis~\cite{yadav2021human}. Being based on finite-state automata, they further support Markov chain-based convergence analysis \cite{zhang2021convergence}.

The contributions of the paper can be summarized as follows. We propose an approach to representing board game configurations in Hex using propositional logic. The expressions describe the presence and absence of pieces with conjunctive clauses (AND-rules). This representation allows us to capture board configurations of arbitrary complexity, inspired by human logical reasoning \cite{human-reasoning}.  Our approach thus falls in the domain of rule-based classifiers, which provide humans with easy-to-understand interpretation of the classifiers' decision making. We further demonstrate how our representation can be aggregated and visualized for global and local interpretation.

\section{Board Game Winner Prediction in Hex}
\subsection{Problem Overview}
Hex is an abstract two-player strategy board game played on an $n \times n$ rhombus board. In brief, the players alternate with placing pieces on empty board slots seeking to form a connection from one edge to the opposite edge. The goal of the game is to form an unbroken chain of pieces from one end of the board to the other. The first player (Black) needs to connect the top edge with the bottom edge, while the second player (White) must form a connection between the left and the right edges.

Predicting the conclusion of the game from any intermediate board state is a challenging task. For example, a 11$\times11$ Hex board has 2.4 $\times$ $10^{56}$ legal positions, and a typical board configuration provides almost three times as many move options compared to Chess\cite{article}. The resulting large branching factor makes it only possible to do relatively shallow game tree searches. As the game approaches the end, predicting the winner through search becomes more feasible, however, from the opening and intermediate board states, only state-of-the-art neural networks are capable of sufficiently accurate winner prediction. Note that unlike Go,  Hex cannot end in a draw, producing a two-class prediction problem.

\subsection{Hex Game Strategies}

To facilitate the interpretability study later in the paper, we here provide a brief overview of basic playing strategies for Hex. In Hex, a good offense is equal to a good defense. If you have completed your connection, it will automatically mean that you have prevented the opponent from completing her connection. And, if you have prevented your opponent's connection, you have, as a consequence, simultaneously completed your own connection. This is why a game of Hex never results in a draw \cite{schachner2019game}.

\begin{figure}[t]
    \centering
    \includegraphics[width=0.7\columnwidth]{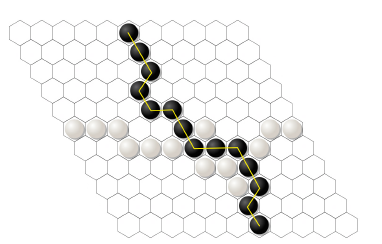}
    \caption{Completed Hex game on an $11 \times 11$ board where Black
is the winner}
    \label{fig:Hex_game01}
\end{figure}
Fig.~\ref{fig:Hex_game01} shows a completed Hex game played on an $11 \times 11$ board. As seen, Black wins the game by forming an unbroken chain from the top edge to the bottom edge. Notice how the continuous chain from top to bottom prevents White from forming a continuous chain from left edge to right edge.

In Hex, the first player always has the advantage. This is because in a finite game, there exists a strategy whereby
a first mover cannot lose~\cite{schwalbe2001zermelo}. The following swap rule balances this bias.

\textbf{Swap Rule:} After the first player (Black) has made its move, the second player can choose either to swap sides, taking the first move as its own, or to respond to the first player, continuing playing as White.

\begin{figure}[t]
    \centering
    \includegraphics[width=0.5\columnwidth]{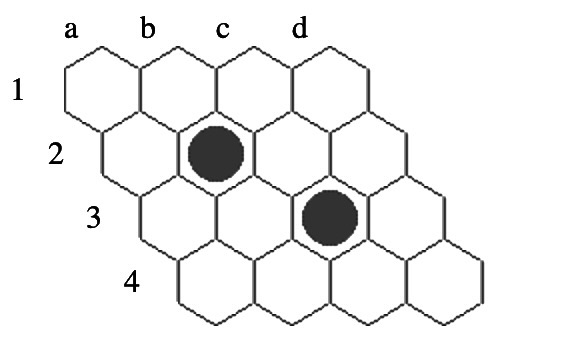}
    \caption{Bridge pattern}
    \label{fig:bridge_pattern}
\end{figure}

Virtual connections and bridge patterns form the basis for playing competitively, illustrated in Fig.~\ref{fig:bridge_pattern}.

\textbf{Virtual Connection}: A virtual connection secures a chain without putting the pieces directly adjacent to each other, lining out a strong chain. The most common type of virtual connection is a bridge pattern.

\begin{figure}[h!]
    \centering
    \includegraphics[width=0.9\columnwidth]{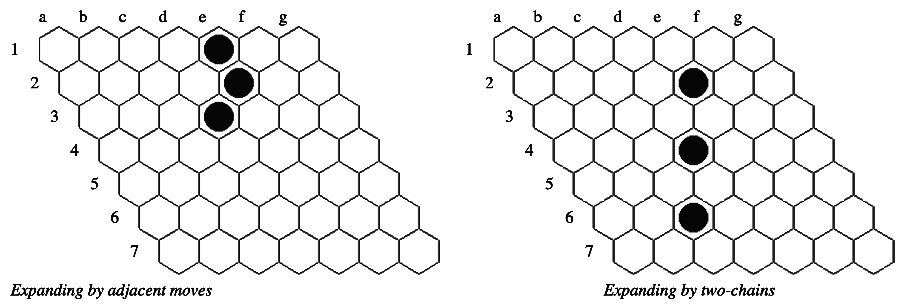}
    \caption{Expansion through bridge pattern}
    \label{fig:bridge_expansion}
\end{figure}

\textbf{Bridge Pattern}: A bridge pattern is created when two pieces of the same color are placed so that they are separated only by two neighbouring empty hex cells. In Fig.~\ref{fig:bridge_expansion}, for example, the link from b2 to c3 is secured by Black, despite the pieces not being adjacent to each other. If White tries to break the link by putting a pieces in b3, for instance, Black can still connect through c2, and vice-versa~\cite{browne2009hex}.

\subsection{Current Approaches}
In 2000, the Hexy~\cite{anshelevich2000game} program won the first Computer Olympiad for Hex. It exploited alpha-beta search and H-search to find and combine virtual connections to form more complex patterns.
Later, MoHex\cite{arneson2009mohex} became champion in the 2009 Hex Computer Olympiad. MoHex is a Monte Carlo tree search (MCTS) based approach. It combines MCTS with H-search and ICA to prune the game tree and speed up the tree search. It further exploits Depth-first Proof Number search to find the best moves in parallel.
MoHex 2.0~\cite{huang2013mohex} is an improvement over MoHex, using the MiniMax algorithm to learn patterns for weighing the MCTS simulations.

From 2009 to 2016, different versions of MoHex won all of the Computer Olympiads. In 2016, however, inspired by the success of AlphaGo~\cite{silver2016mastering}, researchers started applying neural networks to the game of Hex. NeuroHex~\cite{young2016neurohex} applied Deep Q-Learning with experience replay. Another program called MoHexNet~\cite{haywardmohex} used depth-1 trees found by NeuroHEX, and combined the depth-1 trees with MCTS. As a result, MoHexNet won the 2016 Computer Olympiad. Since then, recent work introduced  various ways of improving MoHexNet, leading to MoHex-CNN~\cite{hayward2017hex} and MoHex-3HNN\cite{gao2019hex}, which won the Computer Olympiads in the following years.

Extensive research has been carried out on the policy evaluation methods for predicting next best moves, playing out the entire game with MCTS. While  a MCTS search by itself is interpretable, the neural network-based evaluation of board positions are difficult to interpret due to the lack of transparency of neural networks. In our present paper, we focus on predicting the winner directly, from any board position, without playing out the game. Our goal is to address the lacking transparency of the neural network-based approaches, hindering for instance human-AI collaboration.

Our paper sheds light on this matter by introducing an interpretable method for predicting the winner for $6\times6$ Hex boards. Also, Hex, just like Go, is a pattern oriented game. So this research can be regarded as a step in the direction of developing AI assistance for the game of Go as well.

\section{Interpretable Winner Prediction with Logical Rules}
\label{Tsetlin Machine}

In this section, we present the basics of the Tsetlin Machine, to lay the foundation for our board game winner prediction approach. We further describe our method for representing winning and losing board game configurations.

%

\begin{figure}[b]
  \begin{center}
  \includegraphics[width=\linewidth]{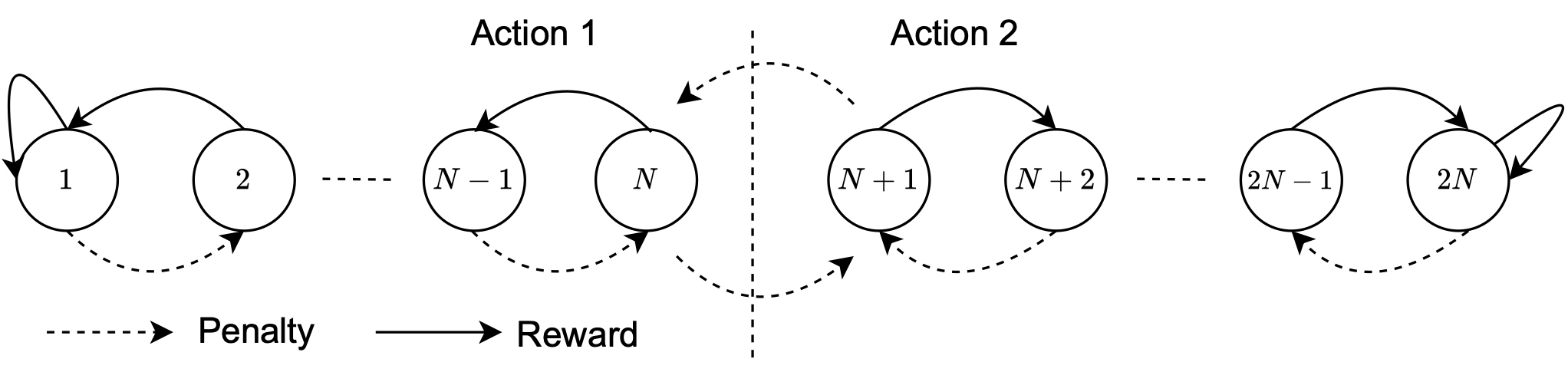}
  \caption{A two-action Tsetlin Automaton with $2N$ states.}\label{figTA}
  \end{center}
\end{figure}

\begin{figure}[ht]
\begin{center}
\centerline{\includegraphics[width=\columnwidth]{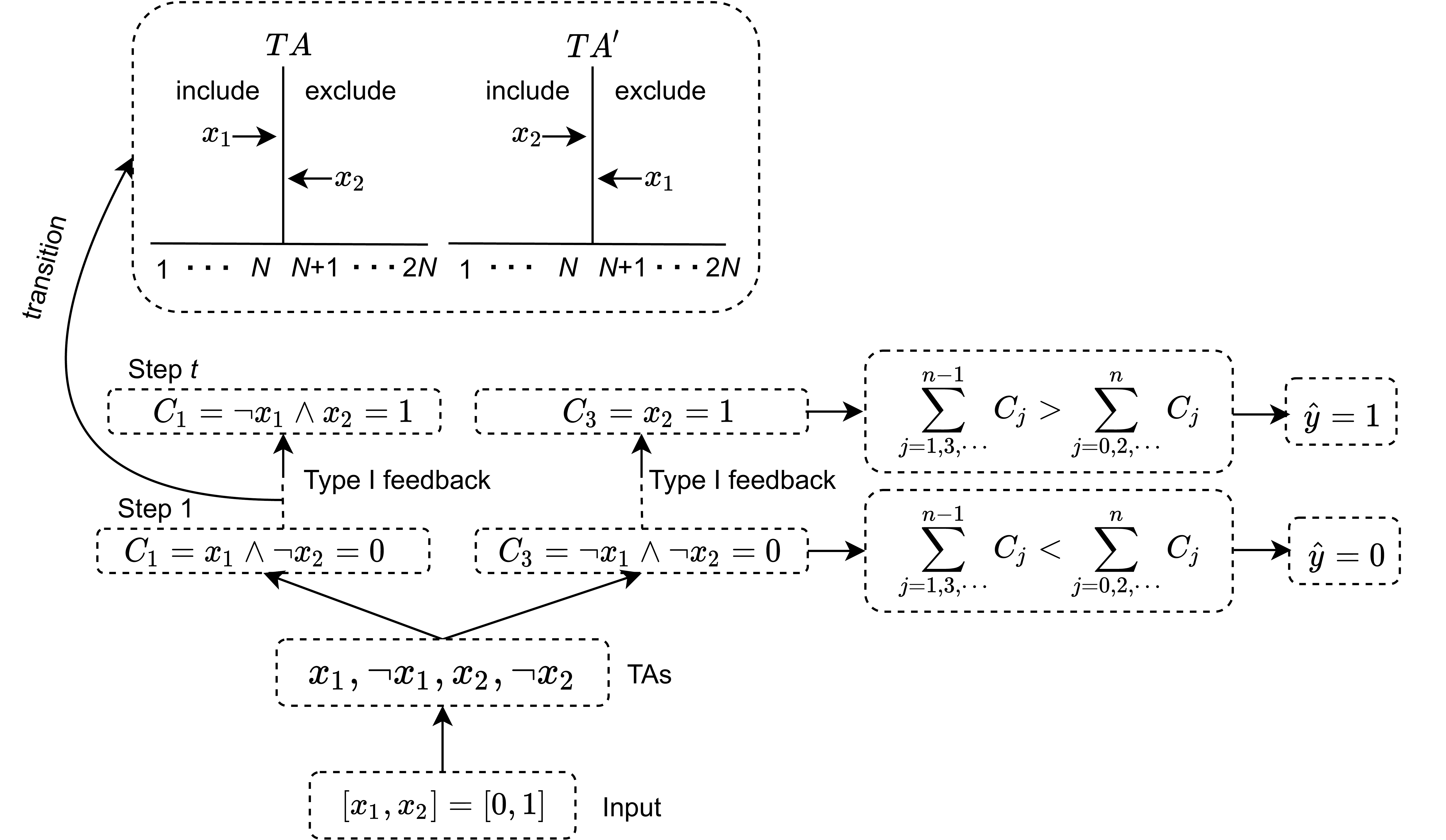}}
\caption{TM learning dynamics for an XOR-gate training sample, with input ($x_1=0, x_2=1$) and output target $y=1$.}
\label{figure:tm_architecture_basic}
\end{center}
\end{figure}

\subsection{Pattern Representation}

A TM in its simplest form takes a feature vector $\mathbf{x} = [x_1, x_2, \ldots, x_o] \in \{0,1\}^o$ of $o$ propositional values as input and assigns the vector a class $\hat{y} \in \{0,1\}$. To minimize classification error, the TM produces $n$ self-contained patterns. The input vector $\mathbf{x}$ first provides the literal set $L = \{l_1, l_2, \ldots, l_{2o}\} = \{x_1, x_2, \ldots, x_{o}, \lnot x_1, \lnot x_2, \ldots, \lnot x_o\}$, consisting of the input features and their negations. By selecting subsets $L_j \subseteq L$ of the literals, the TM can build arbitrarily complex patterns through ANDing the literal selection to form conjunctive clauses:
\begin{equation}
C_j(\mathbf{x})= \bigwedge_{l_k \in L_j} l_k.
\end{equation}
Above, $j \in \{1, 2, \ldots, n\}$ refers to a particular clause $C_j$ and $k \in \{1, 2, \ldots, 2o\}$ refers to a particular literal $l_k$. As an example, the clause $C_j(\mathbf{x}) = x_1 \land \lnot x_2$ consists of the literals $L_j = \{x_1, \lnot x_2\}$ and evaluates to $1$ when $x_1=1$ and $x_2=0$. 

\subsection{Representing Board Game Configurations}

\begin{figure}[H]
    \centering
    \includegraphics[width=0.7\columnwidth]{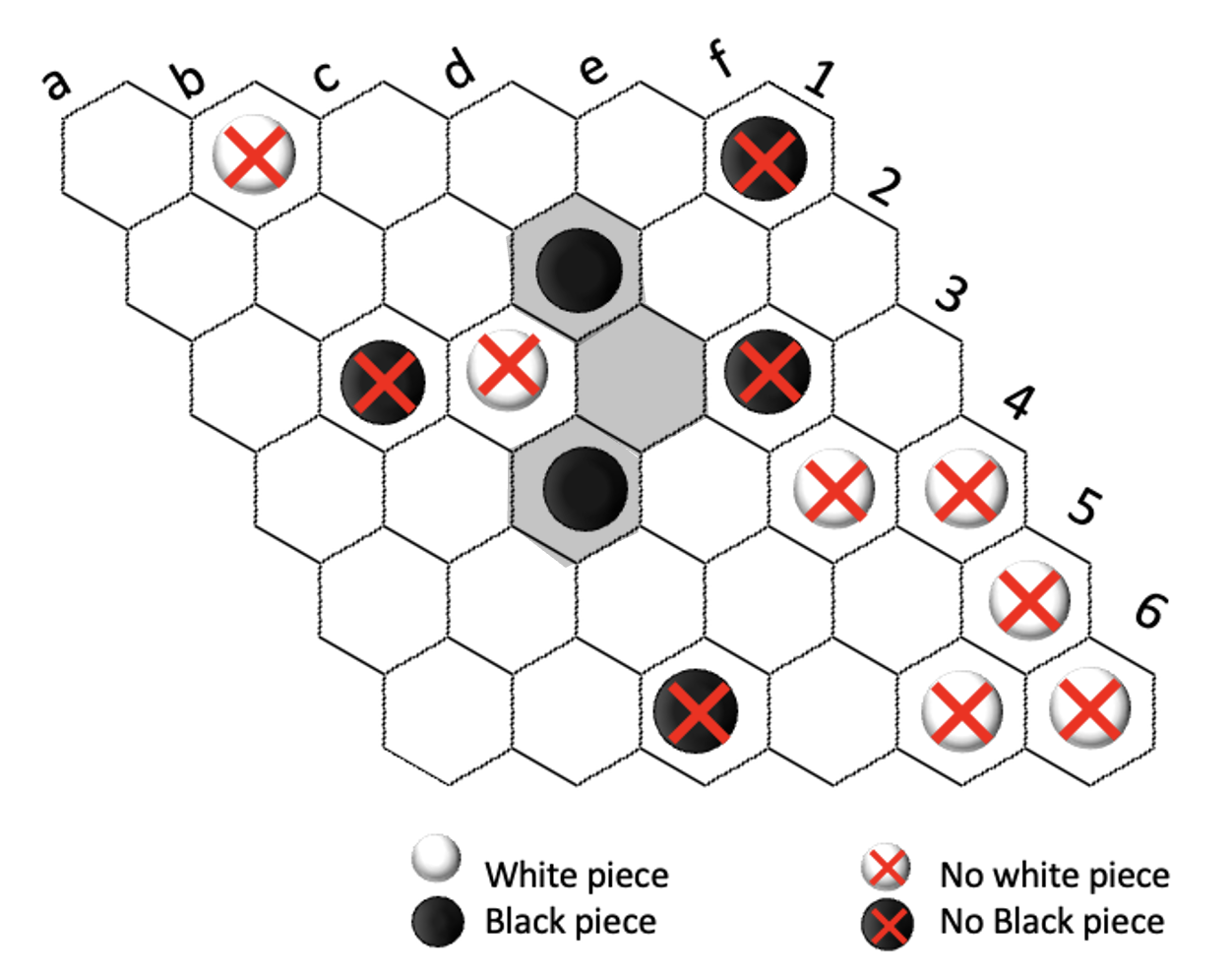}
    \caption{Pattern structure for a sample clause}
    \label{fig:Example_pattern}
\end{figure}

We represent a board game configuration as a clause, with each literal referring to whether a piece of a certain color is present or absent in a specific board location. The following is an example of such a clause, capturing the $6 \times 6$ Hex board pattern visualized in Fig.~\ref{fig:Example_pattern}:\\

\begin{math}
x_{10} \land x_{21} \land \neg x_6 \land \neg x_{14} \land \neg x_{17} \land \neg x_{33} \land \neg x_{38} \land \neg x_{51} \land \neg x_{59} \land \neg x_{60} \land \neg x_{66} \land \neg x_{71} \land \neg x_{72}
\end{math}.
\\
\\
Features $x_1, x_2, \ldots, x_{36}$ refer to the Black pieces, index starting from the top left of the board and increasing along the rows. For instance, literal $x_{10}$ in the clause specifies that there must be a Black piece in position \emph{d2}, while literal $\neg x_6$ signify that a Black piece cannot be located in location \emph{f1}. Correspondingly, features $x_{31}, x_{32}, \ldots, x_{72}$ represent the White pieces. This particular example pattern provides evidence for Black winning because the positions \emph{c4} and \emph{d2} are forming a bridge connection. However, the final prediction is not based on a single individual pattern but by sharing the knowledge of multiple learned patterns.

\subsection{Tsetlin Automata Teams}

The fundamental learning unit of TM is a Tsetlin Automaton(TA). The TM assigns one TA per literal $l_k$ per clause $C_j$ to build the clauses. The TA assigned to literal $l_k$ of clause $C_j$ decides whether $l_k$ is \emph{Excluded} or \emph{Included} in $C_j$. Fig.~\ref{figTA} depicts a two-action TA with $2N$ states.  For states $1$ to $N$, the TA performs action \emph{Exclude} (Action 1), while for states $N + 1$ to $2N$ it performs action \emph{Include} (Action 2). As feedback to the action performed, the environment responds with either a Reward or a Penalty. If the TA receives a Reward, it moves deeper into the side of the action. If it receives a Penalty, it moves towards the middle and eventually switches action.

With $n$ clauses and $2o$ literals, we get $n\times2o$ TAs. We organize the states of these in a $n\times2o$ matrix $A = [a_k^j] \in \{1, 2, \ldots, 2N\}^{n\times2o}$. We will use the function $g(\cdot)$ to map the automaton state $a_k^j$ to Action $0$ (\emph{Exclude}) for states $1$ to $N$ and to Action $1$ (\emph{Include}) for states $N+1$ to~$2N$: $g(a_k^j) = a_k^j > N$.

We can connect the states $a_k^j$ of the TAs assigned to clause $C_j$ with its composition as follows:
\begin{equation}
C_j(\mathbf{x}) = \bigwedge_{l_k \in L_j} l_k = \bigwedge_{k=1}^{2o} \left[g(a_k^j) \Rightarrow l_k\right].
\end{equation}
Here, $l_k$ is one of the literals and $a_k^j$ is the state of its TA in clause $C_j$. The logical \emph{imply} operator~$\Rightarrow$ implements the \emph{Exclude}/\emph{Include} action. That is, the 
\emph{imply} operator is always $1$ if $g(a_k^j)=0$ (\emph{Exclude}), while if $g(a_k^j)=1$ (\emph{Include}) the truth value is decided by the truth value of the literal.

\subsection{Classification}

The odd-numbered half of the clauses vote for class $\hat{y} = 0$ and the even-numbered half vote for $\hat{y} = 1$. Classification is performed as a majority vote:
\begin{equation}
    \hat{y} = 0 \le \sum_{j=1,3,\ldots}^{n-1} \bigwedge_{k=1}^{2o} \left[g(a_k^j) \Rightarrow l_k\right] - \sum_{j=2,4,\ldots}^{n} \bigwedge_{k=1}^{2o} \left[g(a_k^j) \Rightarrow l_k\right]. \label{eqn:prediction}
\end{equation}
As such, the odd-numbered clauses have positive polarity, while the even-numbered ones have negative polarity. As an example, consider the input vector $\mathbf{x} = [0, 1]$ in the lower part of Fig. \ref{figure:tm_architecture_basic}. The figure depicts two clauses of positive polarity, $C_1(\mathbf{x}) = x_1 \land \lnot x_2$ and $C_3(\mathbf{x}) = \lnot x_1 \land \lnot x_2$ (the negative polarity clauses are not shown). Both of the clauses evaluate to zero, leading to class prediction $\hat{y} = 0$.

\begin{table}[t]
\centering
\vskip 0.15in
\begin{center}
\begin{small}
\begin{sc}
\begin{tabular}{l|l|l|l}
    \hline
    \multirow{2}{*}{Input}&Clause & \ \ \ \ \ \ \ 1 & \ \ \ \ \ \ \ 0 \\
    &{Literal} &\ \ 1 \ \ \ \ \ \ 0 &\ \ 1 \ \ \ \ \ \ 0 \\
    \hline
    \multirow{2}{*}{Include Literal}&P(Reward)&$\frac{s-1}{s}$\ \ \ NA & \ \ 0 \ \ \ \ \ \ 0\\ [1mm]
    &P(Inaction)&$\ \ \frac{1}{s}$\ \ \ \ \ NA &$\frac{s-1}{s}$ \ $\frac{s-1}{s}$ \\ [1mm]
    &P(Penalty)& \ \ 0 \ \ \ \ \ NA& $\ \ \frac{1}{s}$ \ \ \ \ \  $\frac{1}{s}$ \\ [1mm]
    \hline
    \multirow{2}{*}{Exclude Literal}&P(Reward)& \ \ 0 \ \ \ \ \ \ $\frac{1}{s}$ & $\ \ \frac{1}{s}$ \ \ \ \ \
    $\frac{1}{s}$ \\ [1mm]
    &P(Inaction)&$ \ \ \frac{1}{s}$\ \ \ \ $\frac{s-1}{s}$  &$\frac{s-1}{s}$ \ $\frac{s-1}{s}$ \\ [1mm]
    &P(Penalty)&$\frac{s-1}{s}$ \ \ \ \ 0& \ \ 0 \ \ \ \ \ \ 0 \\ [1mm]
    \hline
\end{tabular}
\end{sc}
\end{small}
\end{center}
\caption{Type I Feedback}
\label{table:type_i}
\end{table}

\begin{table}[t]
\centering
\vskip 0.15in
\begin{center}
\begin{small}
\begin{sc}

\begin{tabular}{l|l|l|l}
    \hline
    \multirow{2}{*}{Input}&Clause & \ \ \ \ \ \ \ 1 & \ \ \ \ \ \ \ 0 \\
    &{Literal} &\ \ 1 \ \ \ \ \ \ 0 &\ \ 1 \ \ \ \ \ \ 0 \\
    \hline
    \multirow{2}{*}{Include Literal}&P(Reward)&\ \ 0 \ \ \ NA & \ \ 0 \ \ \ \ \ \ 0\\[1mm]
    &P(Inaction)&1.0 \ \  NA &  1.0 \ \ \ 1.0 \\[1mm]
    &P(Penalty)&\ \ 0 \ \ \ NA & \ \ 0 \ \ \ \ \ \ 0\\[1mm]
    \hline
    \multirow{2}{*}{Exclude Literal}&P(Reward)&\ \ 0 \ \ \ \ 0 & \ \ 0 \ \ \ \ \ \ 0\\[1mm]
    &P(Inaction)&1.0 \ \ \ 0 &  1.0 \ \ \ 1.0 \\[1mm]
    &P(Penalty)&\ \ 0 \ \  1.0 & \ \ 0 \ \ \ \ \ \ 0\\[1mm]
    \hline
\end{tabular}
\end{sc}
\end{small}
\end{center}
\caption{Type II Feedback}
\label{table:type_ii}
\end{table}

\subsection{Learning}

The upper part of Fig. \ref{figure:tm_architecture_basic} illustrates learning. A TM learns online, processing one training example $(\mathbf{x}, y)$ at a time. Based on $(\mathbf{x}, y)$, the TM rewards and penalizes its TAs, which amounts to incrementing and decrementing their states. There are two kinds of feedback: Type I Feedback encourages clauses to encode frequently-encountered patterns and Type II Feedback increases the discrimination power of the patterns.

\begin{figure}[bt]
     \centering
         \centering
         \includegraphics[width=0.5\textwidth]{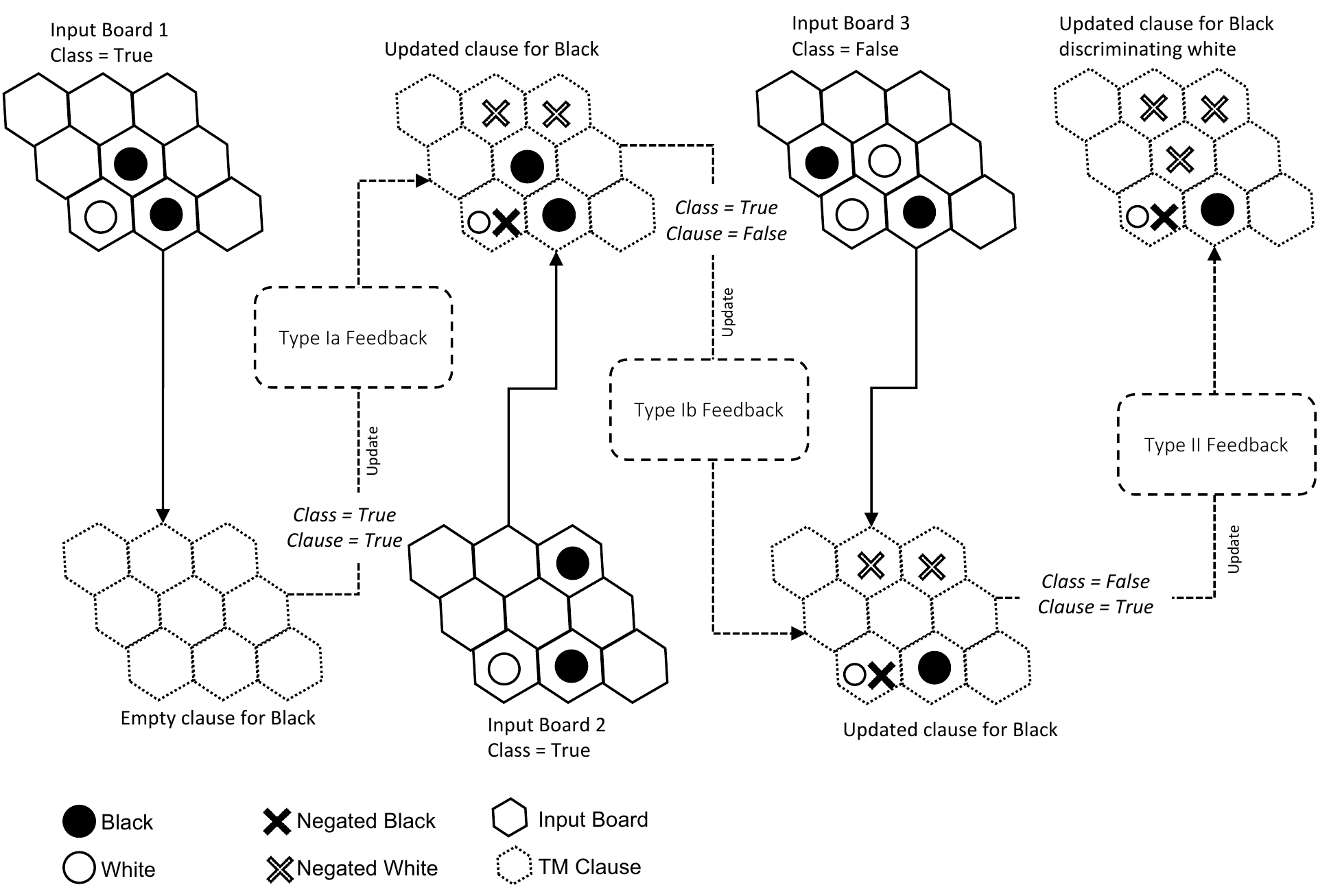}
         \caption{TM learning dynamics for three Hex board positions}
         \label{fig:feedback_visualisation}
\end{figure}  

Type I feedback is given stochastically to clauses with positive polarity when $y=1$  and to clauses with negative polarity when $y=0$. Conversely, Type II Feedback is given stochastically to clauses with positive polarity when $y=0$ and to clauses with negative polarity when $y=1$. The probability of a clause being updated is based on the vote sum $v$: $v = \sum_{j=1,3,\ldots}^{n-1} \bigwedge_{k=1}^{2o} \left[g(a_k^j) \Rightarrow l_k\right] - \sum_{j=2,4,\ldots}^{n} \bigwedge_{k=1}^{2o} \left[g(a_k^j) \Rightarrow l_k\right]$. The voting error is calculated as:
\begin{equation}
\epsilon = \begin{cases}
T-v& y=1\\
T+v& y=0.
\end{cases}
\end{equation}
Here, $T$ is a user-configurable voting margin yielding an ensemble effect. The probability of updating each clause is $P(\mathrm{Feedback}) = \frac{\epsilon}{2T}$.

A random sampling from $P(\mathrm{Feedback})$ decides which clauses to update. The following TA state updates of the chosen clauses' can be formulated as matrix additions, subdividing Type I Feedback into feedback Type Ia and Type Ib:
\begin{equation}
    A^*_{t+1} = A_t + F^{\mathit{II}} + F^{Ia} - F^{Ib}.
    \label{eqn:learning_step_1}
\end{equation}
Here, $A_t = [a^j_k] \in \{1, 2, \ldots, 2N\}^{n \times 2o}$ contains the states of the TAs at time step $t$ and $A^*_{t+1}$ contains the updated state for time step $t+1$ (before clipping). The matrices $F^{\mathit{Ia}} \in \{0,1\}^{n \times 2o}$ and $F^{\mathit{Ib}} \in \{0,1\}^{n \times 2o}$ contains Type I Feedback. A zero-element means no feedback and a one-element means feedback. As shown in Table \ref{table:type_i}, two rules govern Type I feedback:
\begin{itemize}
    \item \textbf{Type Ia Feedback} is given with probability $\frac{s-1}{s}$ whenever both clause and literal are $1$-valued.\footnote{Note that the probability $\frac{s-1}{s}$ is replaced by $1$ when boosting true positives.} It penalizes \emph{Exclude} actions and rewards \emph{Include} actions. The purpose is to remember and refine the patterns manifested in the current input $\mathbf{x}$. This is achieved by increasing selected TA states. The user-configurable parameter $s$ controls pattern frequency, i.e., a higher $s$ produces less frequent patterns.
    \item \textbf{Type Ib Feedback} is given with probability $\frac{1}{s}$ whenever either clause or literal is $0$-valued. This feedback rewards \emph{Exclude} actions and penalizes \emph{Include} actions to coarsen patterns, combating overfitting. Thus, the selected TA states are decreased.
\end{itemize}

The matrix $F^{\mathit{II}} \in \{0, 1\}^{n \times 2o}$ contains Type II Feedback to the TAs, given per Table \ref{table:type_ii}.
\begin{itemize}
\item \textbf{Type II Feedback} penalizes \emph{Exclude} actions to make the clauses more discriminative, combating false positives. That is, if the literal is $0$-valued and the clause is $1$-valued, TA states below $N+1$ are increased. Eventually the clause becomes $0$-valued for that particular input, upon inclusion of the $0$-valued literal.
\end{itemize}

The final updating step for training example  $(\mathbf{x}, y)$ is to clip the state values to make sure that they stay within value $1$ and $2N$:
\begin{equation}
    A_{t+1} = \mathit{clip}\left(A^*_{t+1}, 1, 2N\right). \label{eqn:learning_step_2}
\end{equation}

For example, both of the clauses in Fig.~\ref{figure:tm_architecture_basic} receives Type~I Feedback over several training examples, making them resemble the input associated with $y=1$.

\subsection{Learning Board Configuration Patterns}
\begin{figure}[bt]
    \begin{subfigure}[bt]{0.4\textwidth}
         \centering
         \includegraphics[width=\textwidth]{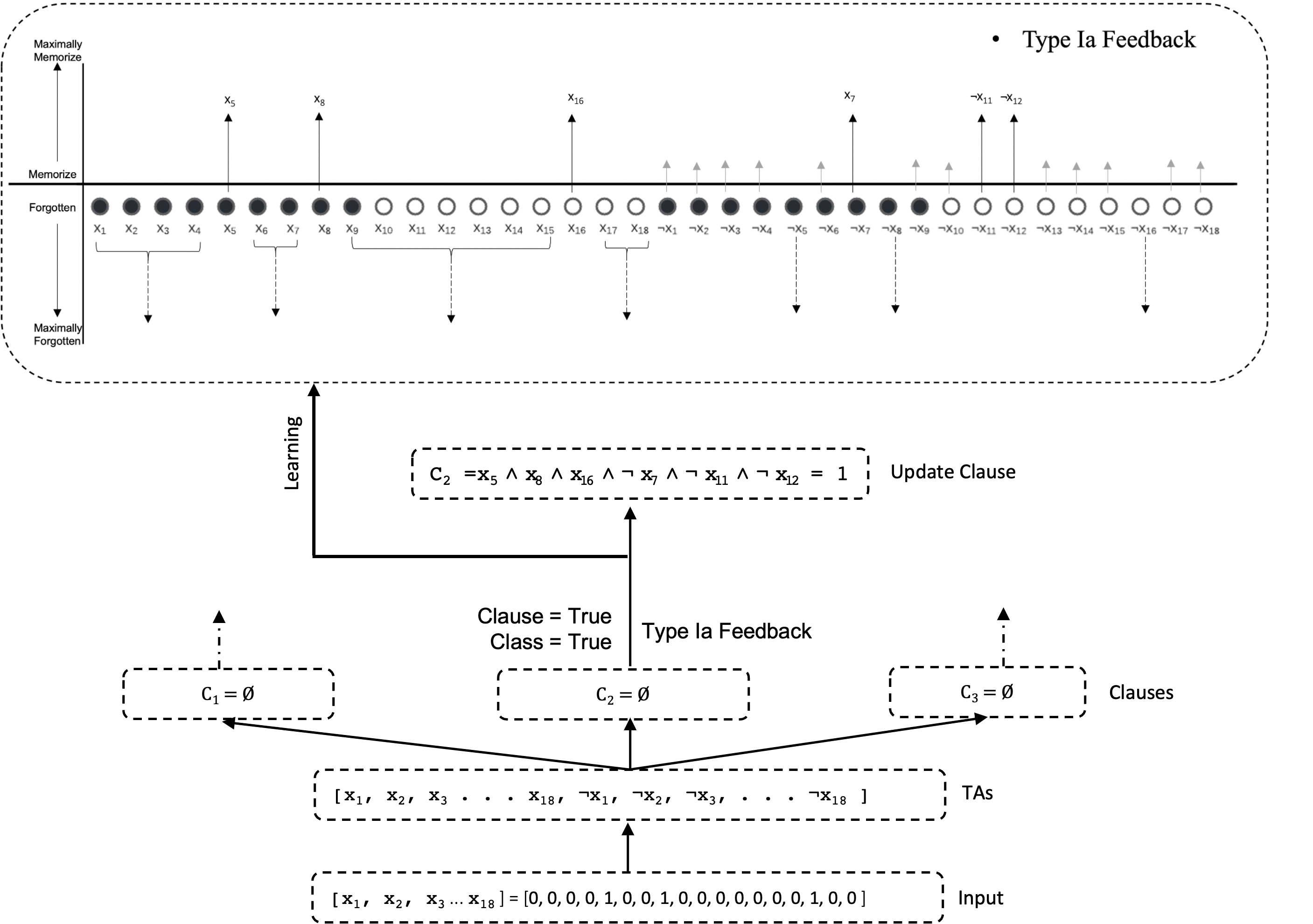}
         \caption{Type Ia Feedback}
         \label{fig:Ia_Feedback}
     \end{subfigure}
     \vskip 0.2in
     \begin{subfigure}[bt]{0.4\textwidth}
         \centering
         \includegraphics[width=\textwidth]{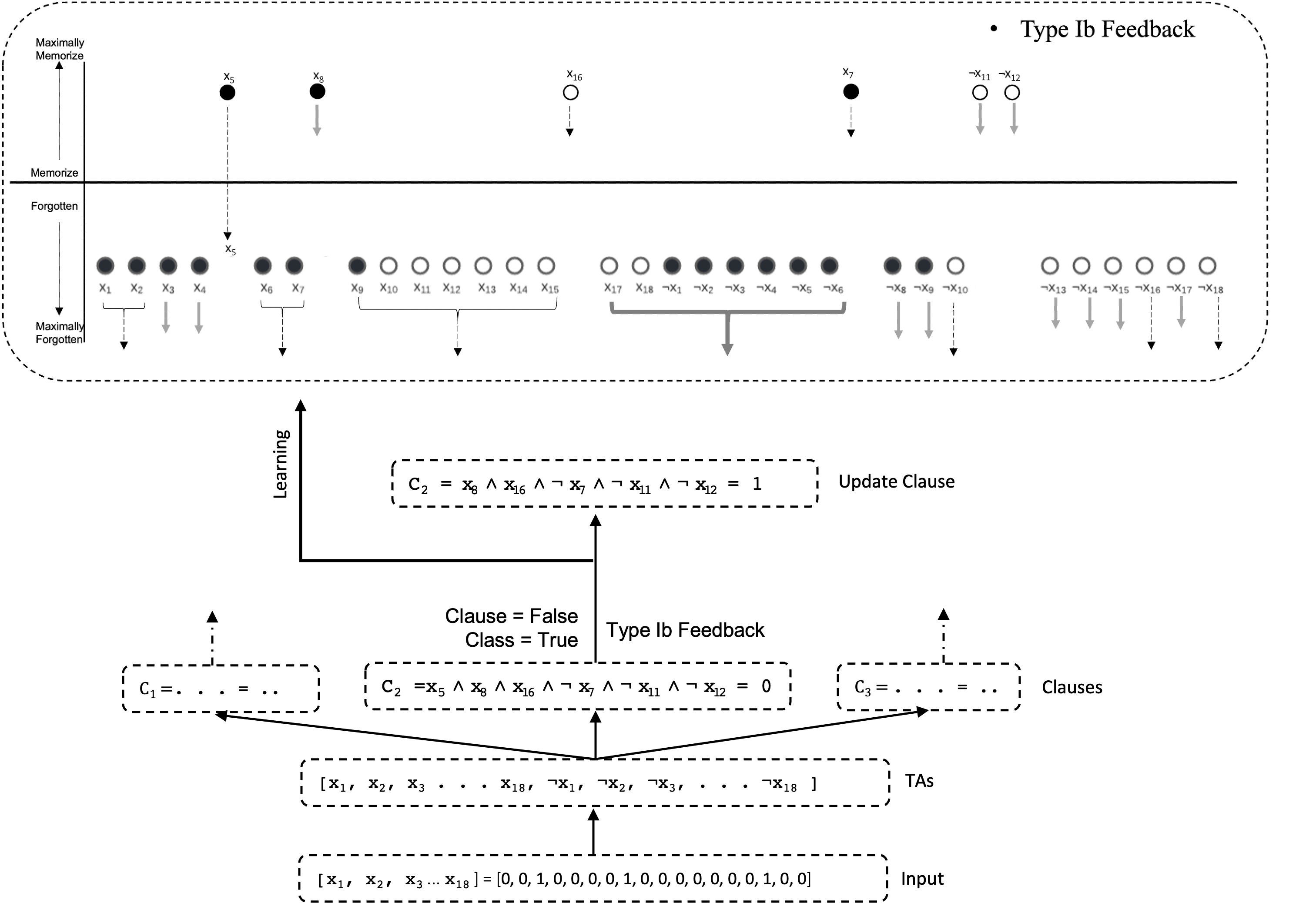}
         \caption{Type Ib Feedback}
         \label{fig:Ib_Feedback}
     \end{subfigure}
     \vskip 0.2in
     \begin{subfigure}[bt]{0.4\textwidth}
         \centering
         \includegraphics[width=\textwidth]{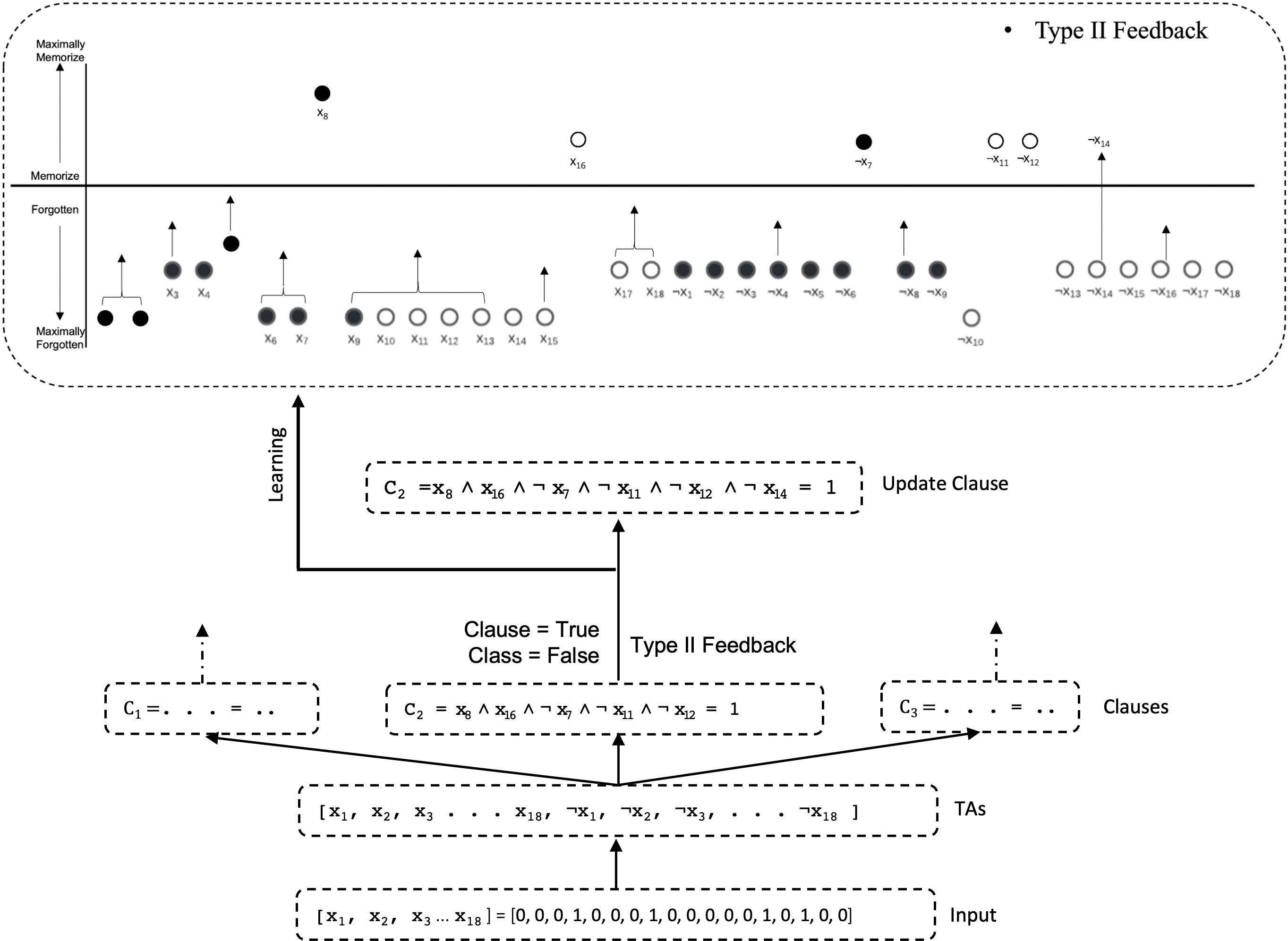}
         \caption{Type II Feedback}
         \label{fig:II_Feedback}
     \end{subfigure}
     
    \caption{Visualization of Tsetlin Automata state updates}
    \label{fig:Learning_Visualization}
\end{figure}
Fig.~\ref{fig:feedback_visualisation} demonstrates how Type~Ia, Type~Ib, and Type~II Feedback operates on board game configurations to update a clause. We use $3\times3$ boards to visualize learning. The starting point is an empty clause, which is updated based on an input board that led to a Black win (upper left). Since the clause is empty, it evaluates to True. Type Ia Feedback then makes the clause resemble the input board by introducing literals from the board, building a pattern for winning positions. The next new winning board configuration given as input does not match the clause, leading to Type Ib Feedback that removes literals from the clause. Finally, a losing board position is given as input. Since the clause evaluates to True, Feedback Type II is given, refining the clause to become more discriminative (upper left). It is the states of the underlying Tsetlin Automata that drive these changes, illustrated by Fig. \ref{fig:Learning_Visualization}).

\section{Empirical Evaluation}
In this section we evaluate our proposed method and demonstrate how to interpret the board configuration patterns produced.

\subsection{Dataset} \label{Empirical Evaluation}
The dataset we use for evaluation consists of $287,000$ game board configurations, which were created through self-play using a modified version of MoHex2.0, with MCTS and data augmentation. The games lasted for up to 28 moves, with the majority of games ending before 20 moves (Fig.~\ref{fig:datastat}). 

Since TM requires input features to be binarized, each board is represented with a $72$ bit vector, as described in the previous section.
\begin{table}[bt]
\centering
\vskip 0.3in
\begin{center}
\begin{small}
\begin{sc}
\renewcommand{\arraystretch}{1.2}
\begin{tabular}{l|l}
                \hline
                Class       &    Game Boards \\
                \hline
                Black       &    \makecell{175968} \\
                White       &    \makecell{111826} \\
                \hline
                Total       &    \makecell{287794} \\
                \hline
\end{tabular}
\end{sc}
\end{small}
\end{center}
\caption{Dataset distribution}
\label{table:type_ii}
\end{table}

\begin{figure}[bt]
     \centering
         \includegraphics[width=0.4\textwidth]{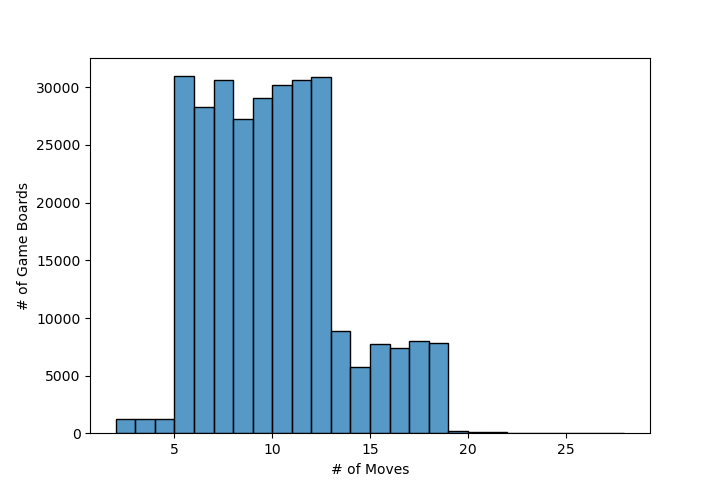}
         \caption{Distribution of games per number of moves played}
         \label{fig:moves_distri}
\end{figure}

\subsection{Results}
Our experimental setup is as follows. As previously explained in Section \ref{Tsetlin Machine}, a TM has two hyper-parameters $T$ and $s$, along with the number of clauses $n$. For our experiments we initialized our TM with $10,000$ clauses and set the other hyperparameter $T$ and $s$ to $8000$ and $100$, respectively. As a general rule, TMs perform well with a $T$ value equal to $80\%$ of the total number of clauses per class,  leaving only $s$ to be tuned. We found an appropriate $s$ value through a limited manual trial and error search, so performance may possibly improve trough a more thorough automated grid search. We used $67\%$ of the data to train the TM for $200$ epochs, and the remaining $33$\% of the data for evaluating prediction accuracy. 

Table \ref{tab:comparison} contains the results of our experiments, including a comparison with six popular machine learning algorithms. For a fair comparison, we used the same training and testing datasets for all of the algorithms. Each algorithm was first trained using default sklearn parameters, and the top-performing ones were then fine-tuned to maximize performance. As seen in the table, we obtained an average testing accuracy of $92.1\%$ with our approach,  outperforming the other interpretable methods like Decision Trees, XGBoost\cite{chen2016xgboost}, and InterpretML\cite{nori2019interpretml} by a wide margin.

Also, our technique performs better than a fine-tuned black-box neural network (NN) configuration. The NN consisted of five layers with Sigmoid activation functions, trained with  the Adam optimizer. We further utilized binary Crossentropy loss function, and exploited $L_2$ regularization and dropout to reduce overfitting. Still, the maximum accuracy we obtained with the NN was about 1.5\% lower than our TM-based approach.

\begin{figure}[t]
     \centering
     
    \begin{subfigure}[b]{0.5\textwidth}
         \centering
         \includegraphics[width=\textwidth]{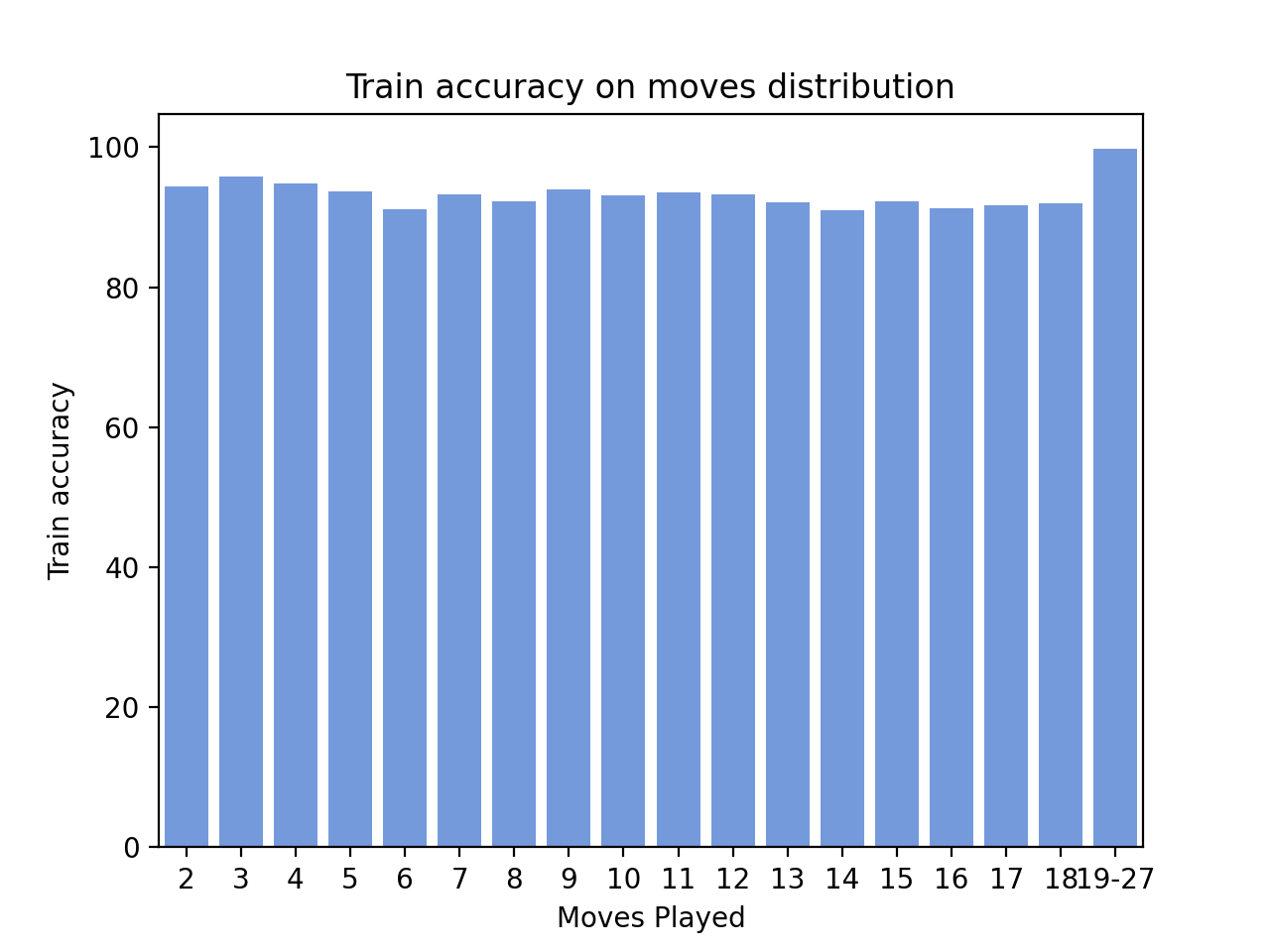}
         \caption{Training accuracy per \#moves played}
         \label{fig:y equals x}
     \end{subfigure}
     \begin{subfigure}[b]{0.5\textwidth}
         \centering
         \includegraphics[width=\textwidth]{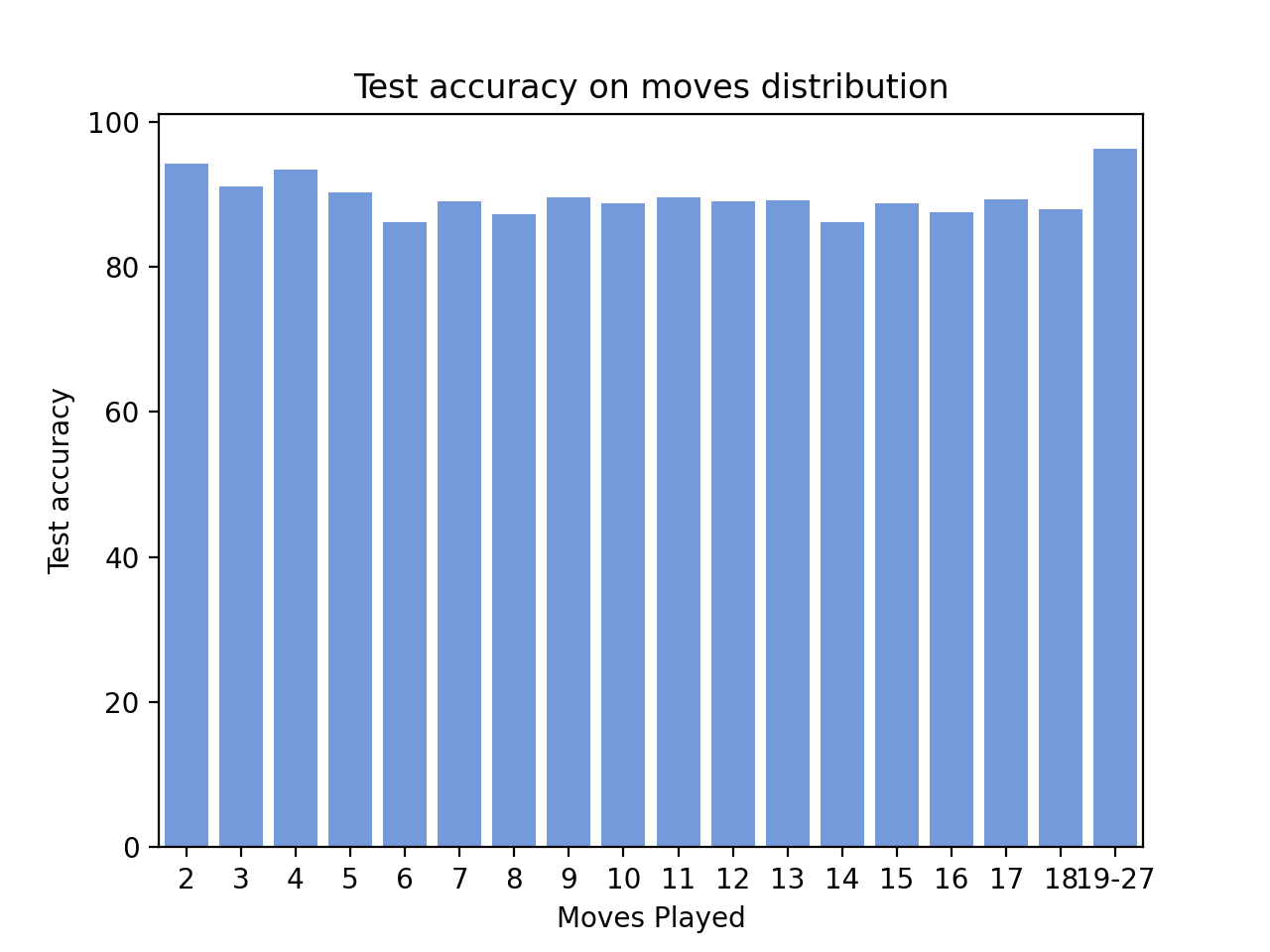}
         \caption{Testing accuracy per \#moves played}
         \label{fig:y equals x}
     \end{subfigure}
    \caption{Accuracy at different stages of a game}
    \label{fig:moves distribution acc}
\end{figure}

To investigate whether accuracy varies depending on the stage of a game (e.g., beginning, middle, or end), we grouped the data according to the number of moves played (Fig.~\ref{fig:moves_distri}). Fig.~\ref{fig:moves distribution acc} summarises how training and testing accuracy varies by the number of moves played. Observe how the performance of TM remains more or less unaffected by number of moves played, indicating a strong ability to predict the winner at all stages of a game. That is, our method is not only able to predict the winner accurately later in a game, when more information is available (intermediate and ending games), it can also predict the winner very early in the game, signifying the importance of the first moves played. 

\begin{figure}[t]
     \centering
    \begin{subfigure}[b]{0.3\textwidth}
         \centering
         \includegraphics[width=\textwidth]{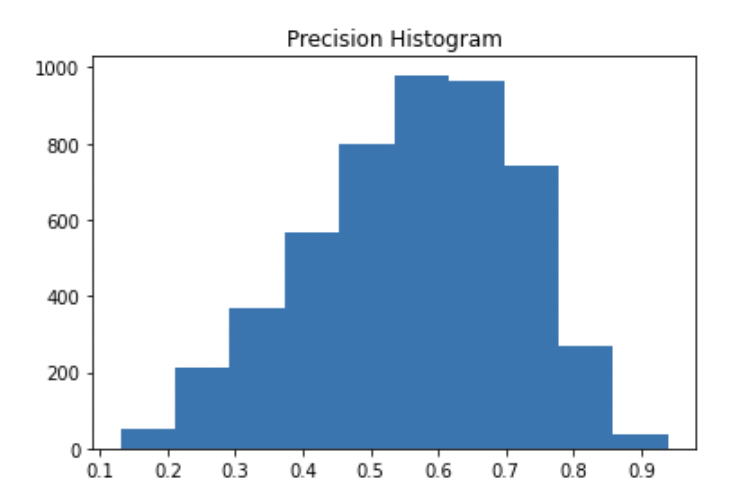}
         \caption{Positive Polarity Clauses}
         \label{fig:y equals x}
     \end{subfigure}
     \begin{subfigure}[b]{0.3\textwidth}
         \centering
         \includegraphics[width=\textwidth]{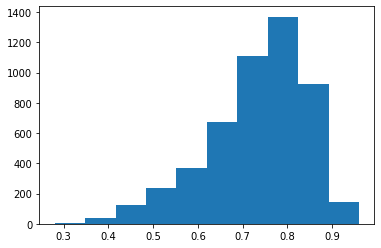}
         \caption{Negative Polarity Clauses}
         \label{fig:y equals x}
     \end{subfigure}
    \caption{Clause precision}
    \label{fig:clause precision}
\end{figure}

Earlier in this paper we discussed that TM learns patterns by forming clauses, so we performed a precision analysis of the clauses. Fig.~\ref{fig:clause precision} illustrates the precision histograms for positive polarity and negative polarity clauses. It can be observed that a majority of the clauses fall in a fairly precise group, which signify that the predictions are made with a higher confidence.


\begin{table}[tb]
\centering
\vskip 0.15in
\begin{center}
\begin{small}
\begin{sc}
\resizebox{\linewidth}{!}{%
\renewcommand{\arraystretch}{1.2}
\begin{tabular}{l | l | l | l}
\hline
Method & Hyperparameter & Training Accuracy & Testing Accuracy \\
\hline

Naïve bayes         &  \makecell{default} & \makecell{72.02} & \makecell{72.02} \\
Decision Tree       & \makecell{max depth=100} & \makecell{100} & \makecell{87.1} \\
K-nearest neighbour & \makecell{k=1} & \makecell{100} & \makecell{84} \\
xgboost             & \makecell{} & \makecell{84.88} & \makecell{88.90} \\
interpretML(EBC)    &  \makecell{default} & \makecell{74.56} & \makecell{74.33} \\
Neural Network      & \makecell{} & \makecell{97.6} & \makecell{90.7} \\
Tsetlin Machine     & \makecell{Clauses = 10000\\ T = 8000\\ S = 100\\max weight=255} & \makecell{95.6}  & \makecell{92.1} \\
\hline
\end{tabular}
}
\end{sc}
\end{small}
\end{center}
\caption{Comparison with other machine learning techniques}
\label{tab:comparison}
\end{table}

\subsection{Interpretation}
We now investigate how to visualize the clauses for global and local interpretation.

\subsubsection{Global Interpretability} Global interpretability~\cite{molnar2020interpretable} describes how a model behaves in general. It takes a holistic view of the features, providing insight into what features are informative overall and how they interact in the model.

\begin{figure}[t]
     \centering
    \begin{subfigure}[t]{0.4\textwidth}
         \centering
         \includegraphics[width=\textwidth]{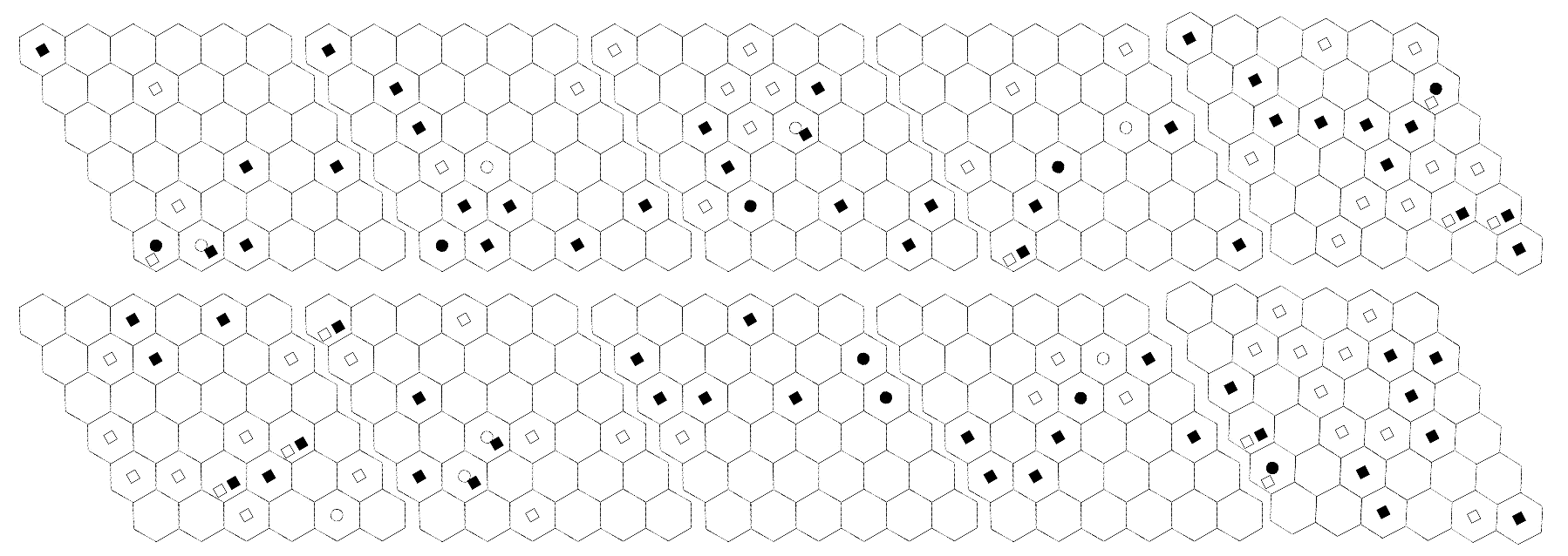}
         \caption{Positive polarity clauses}
         \label{fig:ten_positive_plarity_clauses}
     \end{subfigure}
    \begin{subfigure}[t]{0.4\textwidth}
         \centering
         \includegraphics[width=\textwidth]{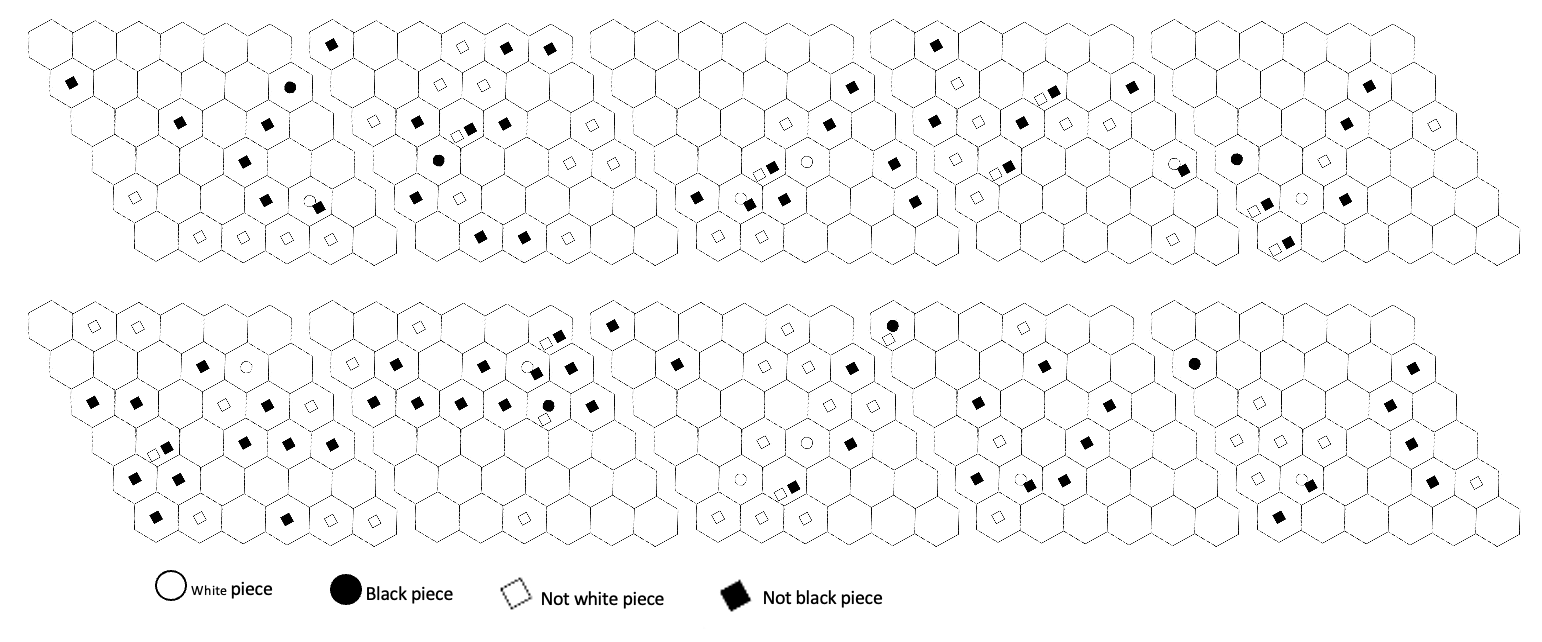}
         \caption{Negative polarity clauses}
         \label{fig:ten_negative_polarity_clauses}
     \end{subfigure}
    \caption{Ten most impactful clauses per polarity}
    \label{fig:clause_visualization}
\end{figure}

In a game of Hex, analysing the interaction of features is crucial for understanding what constitutes a winning pattern. Each clause a TM learns captures one particular feature interaction, as displayed in Fig.~\ref{fig:clause_visualization}. The figure contains the ten most impactful positive polarity clauses (Fig.~\ref{fig:ten_positive_plarity_clauses}) and the ten most impactful negative ones (Fig.~\ref{fig:ten_negative_polarity_clauses}), visualized as board patterns. The positive polarity clauses capture winning patterns for Black, while the negative polarity clauses represent losing patterns.
The clauses $C_j$ have been selected by combining precision and data coverage:
\begin{equation}
Score(j) = Precision(j)^{\alpha}\times Coverage(j)
\end{equation}
calculated from the True Positive (TP), False Negative(FN), and False Positive (FP) rates:
\begin{eqnarray}
Precision =& \frac{TP}{TP + FP}\\
Coverage =& \frac{TP}{TP + FN}. 
\end{eqnarray}
The $\alpha$ parameter is for weighing precision against coverage. That is, a higher $\alpha$ puts more emphasis on clause precision. Since a majority of the clauses provides high coverage, they tend to have relatively lower precision. Therefore, we use $\alpha=10$ to single out clauses with high precision, yet providing reasonable coverage.

\begin{figure}[t]
     \centering
    \begin{subfigure}[t]{0.3\textwidth}
         \centering
         \includegraphics[width=\textwidth]{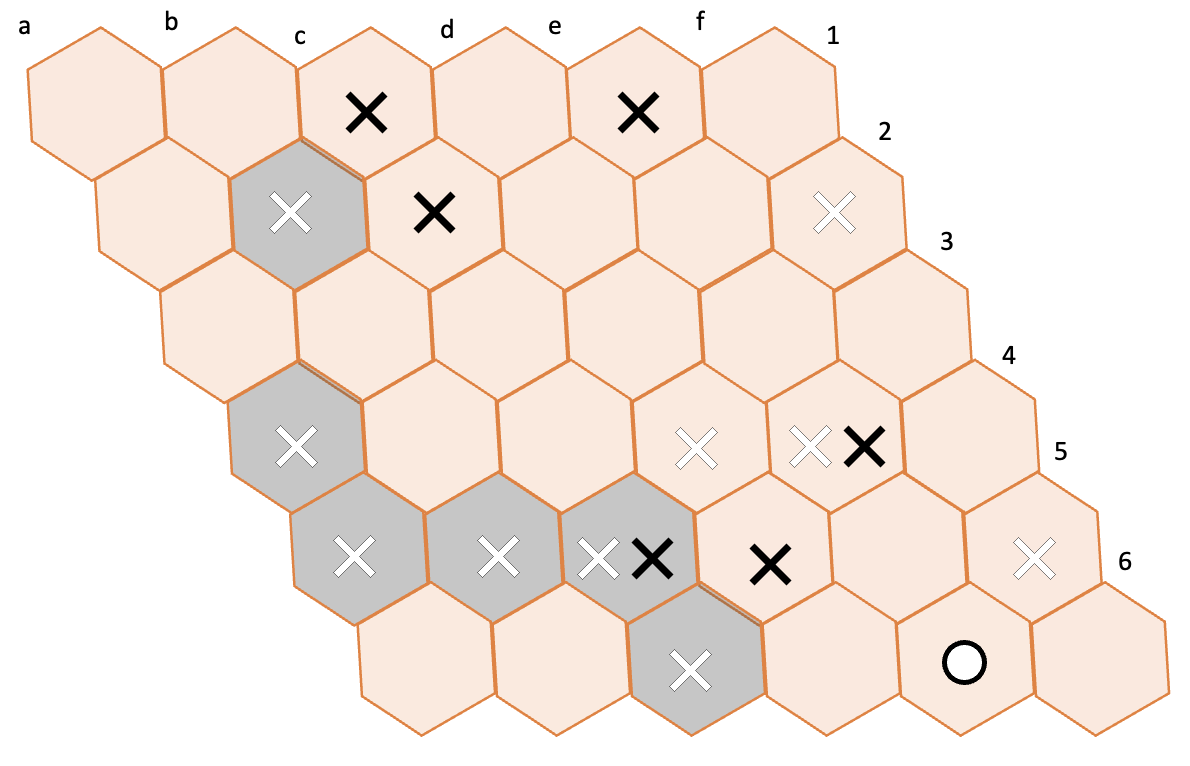}
         \caption{Positive polarity clause}
         \label{fig:GI_PP_interpret}
     \end{subfigure}
    \begin{subfigure}[t]{0.4\textwidth}
         \centering
         \includegraphics[width=\textwidth]{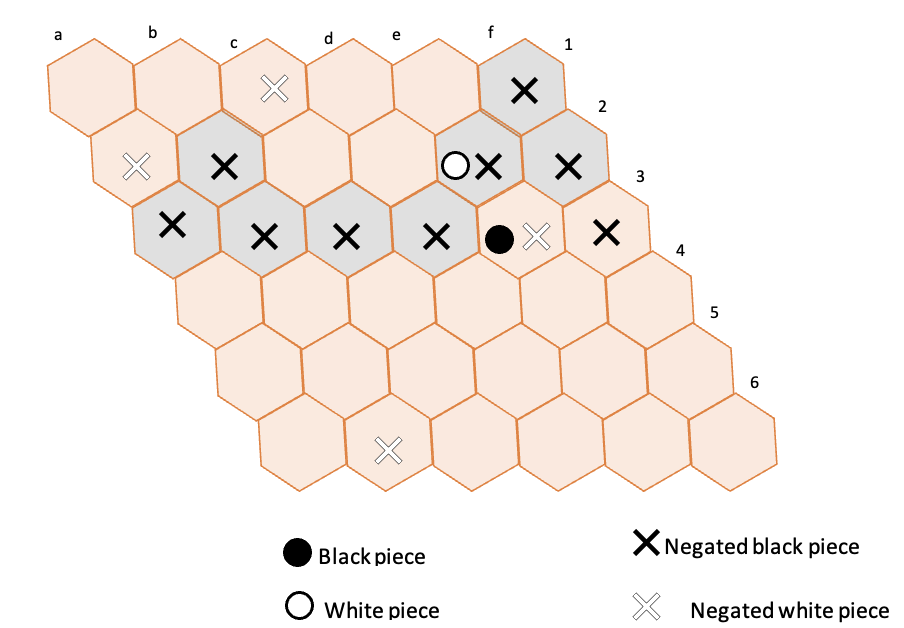}
         \caption{Negative polarity clause}
         \label{fig:GI_NP_interpret}
     \end{subfigure}
    \caption{Example of clause interpretation}
    \label{fig:clause interpret visualization}
\end{figure}

Consider, for example, the negative polarity clause in Fig.~\ref{fig:GI_NP_interpret}. This clause predicts loss for Black due to the string of negated Black pieces from \emph{a3} to \emph{f2}. Hence, the clause states that Black at this point not yet has formed a continuous chain from top to bottom.

Now, observe the positive polarity clause in Fig.~\ref{fig:GI_PP_interpret}. This clause predicts a win for Black. As can be seen from the board, White lacks pieces from \emph{a4} to \emph{c6}. Accordingly, the pattern opens up for a potential chain for Black. The missing white pieces also provide evidence for a possible bridge between \emph{a4} and \emph{b2}, increasing the probability of a Black chain forming before the White one. 

\subsubsection{Local Interpretability}

Local interpretability explains specific data points, as opposed to global interpretation which characterises the entire model at once \cite{molnar2020interpretable}. In our case, global interpretation also forms the basis for local interpretation. That is, by combining the pattern information from the individual clauses that evaluate to True for a specific board configuration, more insights about the board configuration is obtained.

\begin{figure}[t]
     \centering
    \begin{subfigure}[t]{0.4\textwidth}
         \centering
         \includegraphics[width=\textwidth]{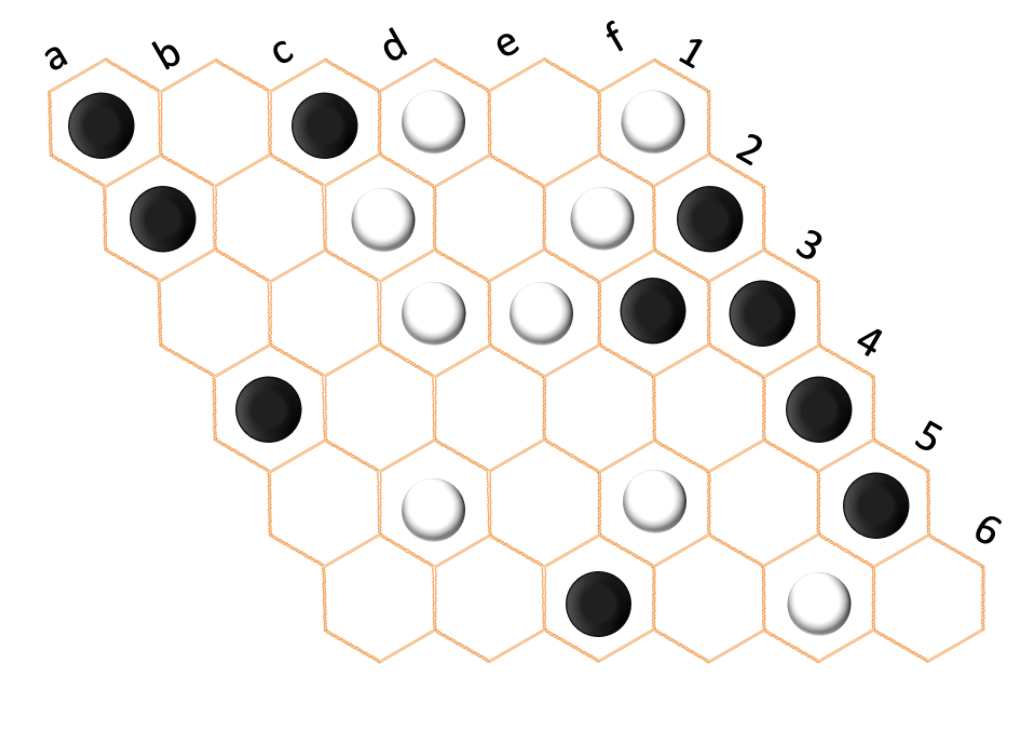}
         \caption{Sample test input board}
         \label{fig:board}
     \end{subfigure}
    \begin{subfigure}[t]{0.4\textwidth}
         \centering
         \includegraphics[width=\textwidth]{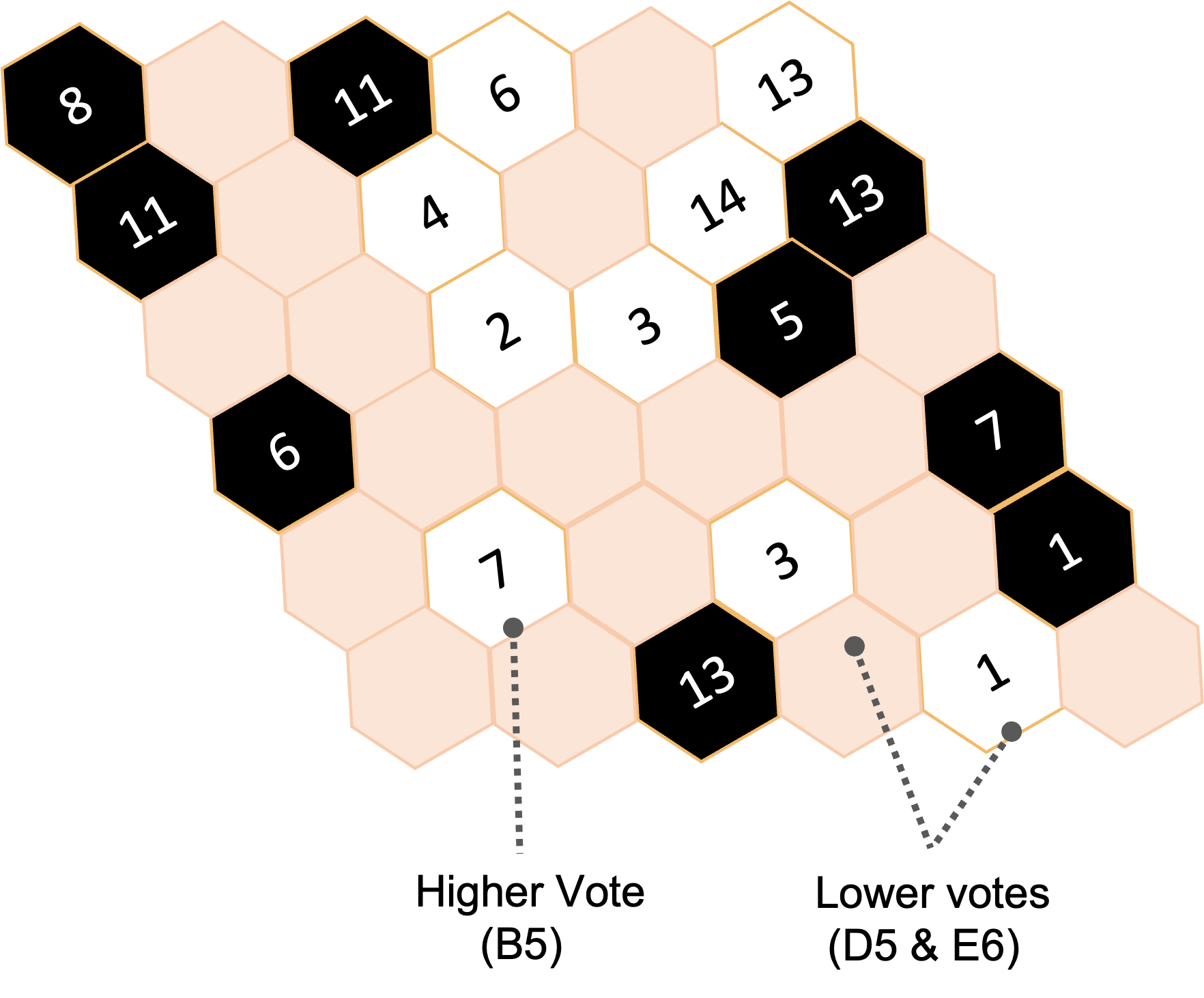}
         \caption{Local interpretation of clauses for board in Fig.~\ref{fig:board} }
         \label{fig:board_interpretation}
     \end{subfigure}
    \caption{Local aggregated interpretation}
    \label{fig:clause_aggregation}
\end{figure}

Since TM clauses work together to form an accurate prediction, it can in some cases be insufficient to interpret a single  clause alone. Instead, we aggregate the information from all the clauses that took part in classifying the input. Then we capture exactly which literals the TM relied on for the given classification.
Consider the example input board in Fig.~\ref{fig:board}. White went on and won the game from this position, as predicted by the TM.
Fig.~\ref{fig:board_interpretation} interprets this TM prediction through aggregating the True clauses. That is, each Hex cell reports the number of times a Black or White piece is assigned to the cell by one of the clauses that are True for the given board position (only considering non-negated features).

We can see that a higher count is given to Hex cell \emph{b5}. The high count captures that \emph{b5} forms a bridge to \emph{c3} as well as a secure edge link to \emph{a5} and \emph{a6}, which completes the continuous chain for White. It is impossible for Black to break this chain. If Black tries to block White by playing \emph{b4}, White can continue the chain by playing \emph{c4}, or the other the way around. In the next step, if Black tries to block White by playing \emph{a6}, White can complete the chain by playing \emph{a5}, and vice-versa.

Also notice the pattern forming in the lower right of Fig.~\ref{fig:board_interpretation}, with a lower count. Even though White can build a chain here as an alternative, it is a weaker option since Black can block White's path from \emph{f1} to \emph{f6} simply by playing \emph{f6}.

In conclusion, this example demonstrate that the TM not only recognizes individual patterns but is also able to aggregate the patterns to assess chance of winning.

\section{Conclusion}

In this paper, we presented an interpretable winner prediction approach for the game of Hex. By using simple AND rules in propositional logic, we can specify strong and weak board configurations. Our empirical results dispute the widespread assumption that one cannot escape trading off interpretability against predictive performance. Despite being inherently interpretable, our approach outperforms several other popular ML methods, achieving an accuracy of $92$\% on test data. Straightforward board configuration patterns enable global and local interpretation, through visualization. By aggregating the active clauses, we can further assess the importance of specific piece formations, revealing the reasons behind our model's predictions.

In our further work we plan to go from winner prediction to predicting the winning move, to solve the game of Hex on larger boards in an interpretable manner.

\bibliography{bibliography}{}

\begin{thebibliography}{10}

\bibitem{abeyrathna2021massively}
Kuruge~Darshana Abeyrathna, Bimal Bhattarai, Morten Goodwin, Saeed~Rahimi
  Gorji, Ole-Christoffer Granmo, Lei Jiao, Rupsa Saha, and Rohan~K Yadav.
\newblock Massively parallel and asynchronous tsetlin machine architecture
  supporting almost constant-time scaling.
\newblock In {\em International Conference on Machine Learning}, pages 10--20.
  PMLR, 2021.

\bibitem{anshelevich2000game}
Vadim~V Anshelevich.
\newblock The game of hex: An automatic theorem proving approach to game
  programming.
\newblock In {\em AAAI/IAAI}, pages 189--194, 2000.

\bibitem{arneson2009mohex}
Broderick Arneson, Ryan Hayward, and Philip Henderson.
\newblock Mohex wins hex tournament.
\newblock {\em ICGA journal}, 32(2):114, 2009.

\bibitem{browne2009hex}
Cameron Browne.
\newblock {\em Hex Strategy: Making the right connections}.
\newblock CRC Press, 2009.

\bibitem{article}
Stephan Chalup, Drew Mellor, and Frances Rosamond.
\newblock The machine intelligence hex project.
\newblock {\em Computer Science Education}, 15, 07 2005.

\bibitem{chen2016xgboost}
Tianqi Chen and Carlos Guestrin.
\newblock Xgboost: A scalable tree boosting system.
\newblock In {\em Proceedings of the 22nd acm sigkdd international conference
  on knowledge discovery and data mining}, pages 785--794, 2016.

\bibitem{dastin2018amazon}
Jeffrey Dastin.
\newblock Amazon scraps secret ai recruiting tool that showed bias against
  women, 2018.

\bibitem{gao2019hex}
Chao Gao, Kei Takada, and Ryan Hayward.
\newblock Hex 2018: Mohex3hnn over deepezo.
\newblock {\em J. Int. Comput. Games Assoc.}, 41(1):39--42, 2019.

\bibitem{articleTM}
Ole{-}Christoffer Granmo.
\newblock The tsetlin machine - {A} game theoretic bandit driven approach to
  optimal pattern recognition with propositional logic.
\newblock {\em CoRR}, abs/1804.01508, 2018.

\bibitem{hayward2017hex}
Ryan Hayward and Noah Weninger.
\newblock Hex 2017: Mohex wins the 11x11 and 13x13 tournaments.
\newblock {\em ICGA Journal}, 39(3-4):222--227, 2017.

\bibitem{haywardmohex}
Ryan Hayward, Noah Weninger, Kenny Young, Kei Takada, and Tianli Zhang.
\newblock Mohex wins 2016 hex 11x11 and 13x13 tournaments.

\bibitem{hipp2011computer}
Jason Hipp, Thomas Flotte, James Monaco, Jerome Cheng, Anant Madabhushi, Yukako
  Yagi, Jaime Rodriguez-Canales, Michael Emmert-Buck, Michael~C Dugan, Stephen
  Hewitt, et~al.
\newblock Computer aided diagnostic tools aim to empower rather than replace
  pathologists: Lessons learned from computational chess.
\newblock {\em Journal of pathology informatics}, 2, 2011.

\bibitem{huang2013mohex}
Shih-Chieh Huang, Broderick Arneson, Ryan~B Hayward, Martin M{\"u}ller, and
  Jakub Pawlewicz.
\newblock Mohex 2.0: a pattern-based mcts hex player.
\newblock In {\em International Conference on Computers and Games}, pages
  60--71. Springer, 2013.

\bibitem{kaur2021trustworthy}
Davinder Kaur, Suleyman Uslu, Arjan Durresi, Sunil Badve, and Murat Dundar.
\newblock Trustworthy explainability acceptance: A new metric to measure the
  trustworthiness of interpretable ai medical diagnostic systems.
\newblock In {\em Conference on Complex, Intelligent, and Software Intensive
  Systems}, pages 35--46. Springer, 2021.

\bibitem{molnar2020interpretable}
Christoph Molnar.
\newblock {\em Interpretable machine learning}.
\newblock Lulu. com, 2020.

\bibitem{nori2019interpretml}
Harsha Nori, Samuel Jenkins, Paul Koch, and Rich Caruana.
\newblock Interpretml: A unified framework for machine learning
  interpretability.
\newblock {\em arXiv preprint arXiv:1909.09223}, 2019.

\bibitem{human-reasoning}
I.~Noveck, R.~B. Lea, George~M. Davidson, and D.~O'brien.
\newblock Human reasoning is both logical and pragmatic.
\newblock {\em Intellectica}, 11:81--109, 1991.

\bibitem{Rudin2019}
Cynthia Rudin.
\newblock {Stop explaining black box machine learning models for high stakes
  decisions and use interpretable models instead}.
\newblock {\em Nature Machine Intelligence}, 1(5):206--215, 2019.

\bibitem{schachner2019game}
Mark Schachner.
\newblock The game of hex: a study in graph theory and algebraic topology.
\newblock {\em Mathematics REU}, 2019.

\bibitem{schwalbe2001zermelo}
Ulrich Schwalbe and Paul Walker.
\newblock Zermelo and the early history of game theory.
\newblock {\em Games and economic behavior}, 34(1):123--137, 2001.

\bibitem{silver2016mastering}
David Silver, Aja Huang, Chris~J Maddison, Arthur Guez, Laurent Sifre, George
  Van Den~Driessche, Julian Schrittwieser, Ioannis Antonoglou, Veda
  Panneershelvam, Marc Lanctot, et~al.
\newblock Mastering the game of go with deep neural networks and tree search.
\newblock {\em nature}, 529(7587):484--489, 2016.

\bibitem{SilHub18General}
David Silver, Thomas Hubert, Julian Schrittwieser, Ioannis Antonoglou, Matthew
  Lai, Arthur Guez, Marc Lanctot, Laurent Sifre, Dharshan Kumaran, Thore
  Graepel, et~al.
\newblock A general reinforcement learning algorithm that masters chess, shogi,
  and go through self-play.
\newblock {\em Science}, 362(6419):1140--1144, 2018.

\bibitem{Tsetlin1961}
Michael~Lvovitch Tsetlin.
\newblock {On behaviour of finite automata in random medium}.
\newblock {\em Avtomat. i Telemekh}, 22(10):1345--1354, 1961.

\bibitem{yadav2021human}
Rohan~Kumar Yadav, Lei Jiao, Ole-Christoffer Granmo, and Morten Goodwin.
\newblock Human-level interpretable learning for aspect-based sentiment
  analysis.
\newblock In {\em The Thirty-Fifth AAAI Conference on Artificial Intelligence
  (AAAI-21). AAAI}, 2021.

\bibitem{young2016neurohex}
Kenny Young, Gautham Vasan, and Ryan Hayward.
\newblock Neurohex: A deep q-learning hex agent.
\newblock In {\em Computer Games}, pages 3--18. Springer, 2016.

\bibitem{zhang2021convergence}
Xuan Zhang, Lei Jiao, Ole-Christoffer Granmo, and Morten Goodwin.
\newblock On the {{Convergence}} of {{Tsetlin Machines}} for the {{IDENTITY}}-
  and {{NOT Operators}}.
\newblock {\em IEEE Transactions on Pattern Analysis and Machine Intelligence},
  2021.

\end{thebibliography}
\bibliographystyle{plain}

\end{document}